# Recognizing Static Signs from the Brazilian Sign Language: Comparing Large-Margin Decision Directed Acyclic Graphs, Voting Support Vector Machines and Artificial Neural Networks

César Roberto de Souza, Ednaldo Brigante Pizzolato, Mauro dos Santos Anjo
Computing Department
Federal University of São Carlos
São Carlos, Brazil
{cesar.souza, ednaldo, mauro_anjo}@dc.ufscar.br

*Abstract*—In this paper, we explore and detail our experiments in a high-dimensionality, multi-class image classification problem often found in the automatic recognition of Sign Languages. Here, our efforts are directed towards comparing the characteristics, advantages and drawbacks of creating and training Support Vector Machines disposed in a Directed Acyclic Graph and Artificial Neural Networks to classify signs from the Brazilian Sign Language (LIBRAS). We explore how the different heuristics, hyperparameters and multi-class decision schemes affect the performance, efficiency and ease of use for each classifier. We provide hyperparameter surface maps capturing accuracy and efficiency, comparisons between DDAGs and 1-vs-1 SVMs, and effects of heuristics when training ANNs with Resilient Backpropagation. We report statistically significant results using Cohen's Kappa statistic for contingency tables.

## I. Introduction

In the last decade, Support Vector Machines (SVMs) have been extensively used, extended and explored with an ever increasing interest, as shown by numerous works of both practical [1,2] and theoretical [3,4,5,6] importance. However, one must not diminish the importance of Artificial Neural Networks (ANNs) nor ignore their many advantageous characteristics already proven in several real world applications. Choosing the best classifier for a given task is thus not as simple as choosing the classifier with higher classification rates. Often, many other factors should be balanced in order to achieve this decision. In this work we provide comparisons between Support Vector Machines (SVMs), Large-Margin Decision Directed Acyclic Graphs (DDAGs) [4] and Artificial Neural Networks (ANNs) in the task of high dimensional classification; both in terms of learning ability, computational costs and ease of use.

This work explores results specific to the task of high dimensional image classification presented in [7]. In the aforementioned work, we created a system for dynamic hand gesture recognition based on a two-stage architecture; developing further upon the work done by [8]. While our previous publication dealt with both static and dynamic signs, this work focuses only on the static image classification problem: the first stage of our previous system.

Our efforts are mainly concentrated towards the Brazilian Sign Language's manual alphabet. In the Brazilian Sign Language, hereinafter LIBRAS, the use of the manual alphabet is only needed in specific occasions. Those occasions include, for example, explicitly spelling the name of a person or a location. Nevertheless, many signs from the manual alphabet are also subcomponents of more elaborated signs and are thus of practical interest as we head towards a fully functional recognition system.

In [8], we considered a classification scheme for dynamic signs based on a two stage architecture. The first stage was responsible for labeling still images (frames) from an image stream into alphabet letters with ANNs. The second stage, in turn, would receive this sequence of coded letters and feed it into a bank of Hidden Markov Models (HMMs), performing classification by the maximum likelihood decision rule. With this architecture, we achieved a system for fingerspelling recognition with promising results.

Furthermore, in [7], we replaced each stage of this architecture with SVMs disposed in large-margin DDAGs. We also replaced the HMM-based classifier with its discriminative counterpart, given by Hidden Conditional Random Fields (HCRFs). Our results have shown how the proper choice of a sequence classifier could play a much bigger impact in the overall performance of the system than would any of the considered choices for frame classifiers. Still, even if overall results seemed not to be affected by the choice of static gesture classifier, the training costs and evaluation speeds associated with those could be of major concern in the applicability of gesture recognition systems in resource-constrained environments, such as within limited processing power, memory capacity, or both. The goal of this paper is thus to address those concerns with an overview of the hyperparameter surfaces for those classifiers both in terms of accuracy and efficiency, discussing their drawbacks, advantages and alternatives.

This paper is organized as follows. After this first introduction, section 2 gives a list of related works, raising some points of interest and discussions. Section 3 gives an overview of the gesture recognition field, its motivation and a brief literature review. Section 4 presents the methods, models and tools used in this work. In section 5 we detail our experiments with fingerspelling recognition, presenting their results in section 6. Next, we conclude our work, giving final considerations and indicating further works in section 7.

## II. Gesture Recognition

Following a comprehensive survey conducted by Mitra and Acharya [9], gesture recognition methods have been traditionally divided into two main categories: Device-based [10] and vision-based [11,12,13,14]. Device-based approaches often constrain the users to wear a tracking device, such as a tracking glove, resulting in less natural interaction. Vision based approaches, on the other hand, free the user from wearing possibly movement limiting and expensive devices. This paper, thus, deals only with vision-based approaches.

Gestures can be either static or dynamic. Static gestures, often called *poses*, are still configurations performed by the user, passive to be registered in a single still image. Dynamic gestures, in turn, vary on time, and have to be captured as a sequence of still images, such as image streams. Often, gestures have both elements, such as in the case of sign languages [9]. In this paper, we will be specifically covering static gesture signs.

## III. Models and Tools

### A. Artificial Neural Networks

As the name implies, at their creation Artificial Neural Networks (ANNs) had a strong biologic inspiration. However, despite their biological origins, ANNs can be seen as a simple function $f: \mathbb{R}^n \to \mathbb{Y}$ mapping a given input $x \in \mathbb{R}^n$ to a corresponding output $y \in \mathbb{Y}$. The output vectors $y = \langle y_1, \ldots, y_m \rangle$ are also restricted to a specific subset of $\mathbb{R}^n$. Each $y_i \in y$ is restricted to a particular range according to the choice of activation function for the output neurons. In the case of a sigmoid activation function, this range is $[0; 1]$; in case of a bipolar sigmoid function, it is $[-1; 1]$.

From a learning perspective, each function (network) $f$ can be said to belong to a class of functions $\mathcal{H}$ sharing a particular form dictated by a choice of architecture and activation functions. Those functions are also parameterized with possible weight vectors $\boldsymbol{\theta} \in \mathbb{R}^w$, in which $w$ is the total number of weight parameters in the network. Since ANNs can be seen as standard mathematical functions, the learning problem can be cast as a standard optimization problem, in which one would like to minimize a divergence, in some sense, between the network outputs $\hat{y}$ and the desired answers $y$. One possible way to achieve this minimization is through the minimization of the error gradient; and a promising method to achieve this is the Resilient Back-propagation algorithm (Rprop) [15,16].

The Rprop algorithm is one of the fastest methods for gradient learning restricted solely to first-order information. Its basic operational principle is to eliminate the (possibly bad) influence of the gradient magnitude in the optimization step. Unlike other gradient based methods, such as Gradient Descent, in which the step size is always proportional to the gradient vector, Rprop takes into account only the direction of the gradient, making it a local adaptive learning algorithm. Because Rprop relies only in first-order information, it is not required to compute (and hence store) the Hessian matrix of second derivatives for the learning problem, making it especially suitable for high dimensional problems.

### B. Support Vector Machines

Unlike ANNs, SVMs seem not to suffer from the curse of dimensionality (although the validity of this claim is sometimes disputed, e.g. [17]). Nevertheless, SVMs have shown great performance in many real-world problems [1,2,11], including high dimensionality [5] and large-scale [3] ones. Interestingly enough, SVMs had their initial roots in the early Perceptron algorithm for learning artificial neurons. The Perceptron algorithm tries to find a hyperplane separating the data, whose decision function is given by

$$h(x) = sgn(\boldsymbol{\theta} \cdot x + b) \underset{\omega_2}{\overset{\omega_1}{\lessgtr}} 0 \quad (1)$$

with $sgn(0) = 1$. Cortes and Vapnik [18,19] proposed learning a separating hyperplane using an approximate version of the Structural Risk Minimization principle: minimizing the structural risk through maximization of the classification margin, while at same time enforcing capacity control by controlling the margin's width. Under those circumstances, the problem can be stated as a constrained optimization problem in the form

$$\min_{w,b,\xi} \frac{1}{2} \|\boldsymbol{\theta}\|^2 + C \left( \sum_{i=1}^{n} \xi_i \right) \quad (2)$$

subject to $y_i(\boldsymbol{\theta} \cdot x_i + b) \geq 1 - \xi_i$ in which $\xi_i \geq 0$ are slack variables and $C$ is a regularization term imposing a weight to the training set error minimization in contraposition to minimizing model complexity. A large value for $C$ would increase the variance of the model, risking to overfit. A small $C$ would in turn lead to possible underfitting. Adding Lagrange multipliers $a_i$, differentiating in respect to $\boldsymbol{\theta}$, $\xi$ and $b$, and imposing stationarity, one arrives at the dual form of the optimization problem given by

$$\max_{\alpha} \sum_{i=1}^{n} \alpha_i - \frac{1}{2} \sum_{i,j}^{n} a_i a_j y_i y_j (x_i \cdot x_j) \quad (3)$$

subject to the bound constraints $0 \leq \alpha_i \leq C, \forall i = 1, \ldots, n$ and $\sum_{i=1}^{n} \alpha_i y_i = 0$. For the interested reader, a much more thorough derivation can be found in [20]. Moreover, to make the classifier nonlinear, one can take advantage of Cover's theorem [21] and consider a nonlinear transformation $\varphi(\cdot): \mathbb{R}^n \to \mathcal{F}$ such that, when applied to input vectors $x_i \in \mathbb{R}^n$, returns their projection in a high-dimensionality feature space $\mathcal{F}$. Considering $\varphi$, one can now write the classifier as

$$h(x) = sgn\left( \sum_{x_j \in SV} \alpha_j y_j \langle \varphi(x_j), \varphi(x) \rangle + b \right). \quad (4)$$

Since the decision function can now be expressed solely in terms of inner products in a feature space, this implies there is no need to compute $\varphi$ explicitly anymore. Those inner products can instead be computed through a Mercer's kernel function [22] of the form

$$k(x, z) = \langle \varphi(x), \varphi(z) \rangle. \qquad (5)$$

Finally, since $\varphi$ does not have to be computed, the feature space $\mathcal{F}$ can also have an arbitrarily high dimensionality, even becoming infinite dimensional, such as in the case of a Gaussian kernel function.

*1) Multiclass classification approaches*

The SVM has been originally conceived as binary-only classifier. Thus it can only decide between two classes at a time. Several approaches have been proposed to generalize SVMs to multiclass problems; one of the most promising being the Large-Margin DDAG [4].

Two of the most common multi-class approaches are the *one-vs-one* and *one-vs-all* classification strategies. For a decision problem over $c$ classes, the *one-vs-all* demands the creation of $c$ classifiers, each trained to distinguish one class from the others. Those classifiers work on binary sub-problems in which instances from a class receives positive labels and all others instances are set to negative labels. In the *one-vs-one (1-vs-1)*, the original multi-class problem is divided into $c(c-1)/2$ sub-problems considering only two classes at a time, and the final decision for a class is obtained through voting. Despite looking more intensive, this method is often cheaper as each sub-problem is typically much smaller than the original problem. This, in turn, may cause a variance increase [23], resulting in an increased chance of *overfitting*. This also leaves the problem of evaluating an increased number of machines for every new instance undergoing classification — which could easily become troublesome or prohibitive in time sensitive applications.

The DDAG, in turn, generalizes the notion of Decision Trees (DTs), allowing for undirected cycles in the reasoning process. The decision can be seen as a process of sequential elimination, in which, at each round, a class is tested against any of the others and the losing class is removed from next decisions. Because each class is eliminated sequentially, the use of DDAGs allows one to conciliate the faster training times of the *1-vs-1* strategy with evaluation speed linear to the number of classes $c$. For $c$ classes, only $(c-1)$ machines are required to be evaluated during runtime. The authors of [4] also have presented strong bounds on the generalization error and VC dimension for those classifiers.

## IV. Experiments

### A. Dataset

The data used in this study had been gathered as part of a previous work [8]. The whole dataset contains static gestures gathered from 45 subjects who articulated 27 signs from the LIBRAS manual alphabet, registered by a single camera in a controlled environment.

For the performed experiments, a subset of 16,200 static gesture samples had been randomly selected from the original static sign set, with spurious or corrupted samples removed. Half of those samples were separated for testing purposes and the other half for training and validation. Hands were located from the still images using Otsu threshold with subsequent cropping and centering. The images were then dimensioned to 32x32 grayscale windows, forming vectors in $\mathbb{R}^{1024}$. Albeit not an optimal representation, this high-dimensionality approach has been done on purpose to study the behavior of the different classifiers without considering further prior information into the problem, such as in the form of more elaborated features or specialized kernels.

### B. Support Vector Machines

Evaluation speed can become a significant drawback when comparing SVMs against other classification methods. In their default formulation, there are no upper bounds on the number of support vectors (SVs) required to solve a given problem. Since the number of SVs often grows linearly with the number of data samples in a training set, this soon becomes problematic. We conducted an experiment in order to measure impact of this problem and check the feasibility of the DDAG decision [4] in overcoming this limitation.

The experiment considered three possible choices of kernel functions; the Gaussian, Quadratic and Linear. For the Gaussian kernel, a coarse-to-fine grid search was conducted in the hyperparameter space in order to find a optimum $\sigma^2$. For each trained machine, we evaluated the testing dataset twice: at first using the *1-vs-1* voting scheme, then with the DDAG decision. We have annotated the performance of the classifiers, measured in terms of Cohen's kappa ($\kappa$), the total number of unique support vectors needed in the voting scheme and the average number of vector evaluations in the DDAG decision path. As linear machines can also be written in a compact form, for linear machines we considered the number of machine evaluations instead of vector evaluations. All SVMs have been learned using the Sequential Minimal Optimization (SMO) algorithm [24]. Initial feasible points for the Gaussian kernel have been identified using the heuristic value for $\sigma$ suggested in presented in [25], based on the inter-quartile range of the norm statistics for the training dataset.

### C. Artificial Neural Networks

If evaluation speed is a concern for SVMs, training behavior is perhaps one of the biggest concerns for ANNs. As their optimization algorithms often have to deal with multiple local minima, the use of heuristics and other learning control mechanisms become almost mandatory. The goal of this experiment was to measure the impact of initialization heuristics, specifically of the Nguyen-Widrow method, in high dimensionality problems. Feed-forward multilayer networks with a single hidden layer have been created with a varying number of hidden nodes in an attempt to enforce capacity control. All networks have been trained until convergence of the mean squared training error using the Rprop algorithm.

## V. RESULTS

Starting with the Gaussian-kernel vector machine, we found a behavior similar to the one described in [6]. More Specifically, we found out that $C$ did not influence the performance of the classifier as much as would a proper choice for $\sigma^2$. Both the $\kappa$ statistic and the number of support vectors (SVs) for each resulting classifier have been found to be mostly dependent on $\sigma^2$ rather than $C$.

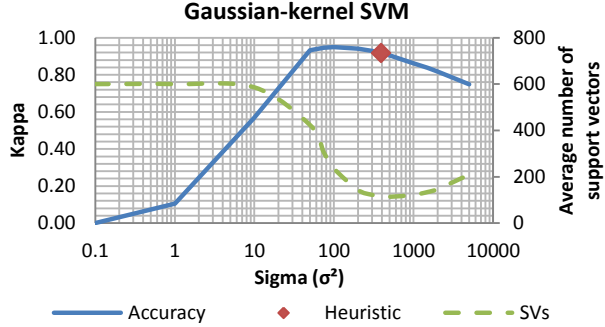

Figure 1. Cuts of the Grid-Search procedure for fixed C.

Figure 1 (above) shows a cut of the grid-search procedure with $C = 1$ and varying $\sigma^2$, alongside with the average number of SVs in the machines. It should be worth to point out that the heuristic choice for $\sigma^2$ as proposed by [25] not only resulted in overall good performance, but also resulted in less SVs. Although not leading to the best possible $\kappa$, the heuristic provided a good balance between accuracy and efficiency for the experimented classifiers.

The hyperparameter surfaces, both in terms of Kappa and sparseness, are shown on Figure 2. From those graphs we can observe an almost direct relationship between the number of SVs and generalization performance; more specifically, there is an apparent inverse correlation between the number of SVs and $\kappa$. This is intuitive, since a higher number of SVs could possibly lead to increased overfitting and therefore to degraded performance.

In this specific problem, $C$ was shown to had little to no influence in the number of SVs, except for high values of $\sigma^2$. In this case, an increase in $C$ seems to counterbalance an increase in $\sigma^2$, leading again to the sparseness plateau centered on the heuristic line ($H \cong 391.52$). The graphs also do not show values for $C$ higher than $10^8$ as those values failed to converge. This could be understandable, as for higher values of $C$ the soft-margin SVM approximates the hard-margin decision, reducing its ability to cope with misclassifications in the training set.

After the Gaussian, the next step was to measure the Linear and Quadratic kernels. Before we proceed, it should be worth to point that the problem we are exploring already has a large number of dimensions. The Linear kernel was thus already expected to produce good results, and this suspicion was confirmed in Figure 3. Both kernels, Linear and Quadratic, produced overall good results. And albeit not depicted, results have shown no significantly different results from using homogeneous or inhomogeneous kernels.

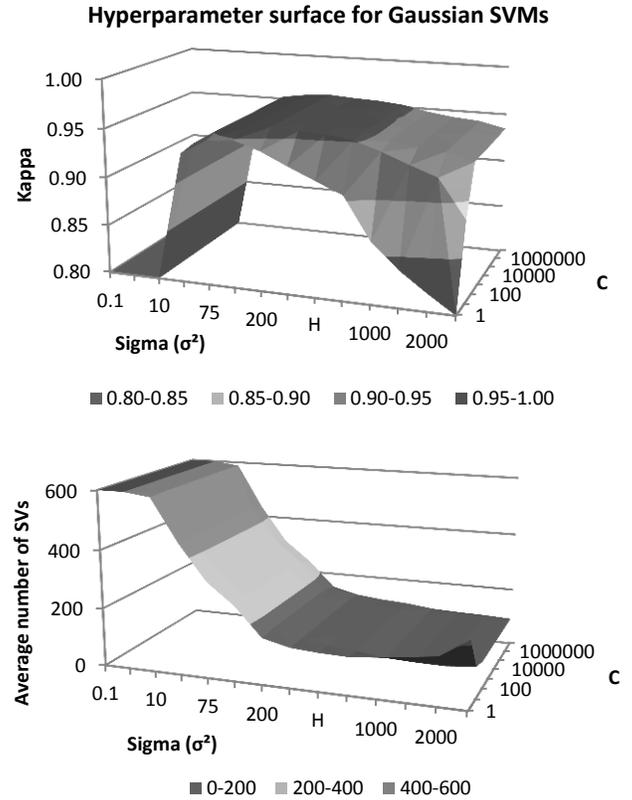

Figure 2. Surface plots for the performance (Kappa) and sparseness (average number of support vectors) of Gaussian-kernel SVMs.

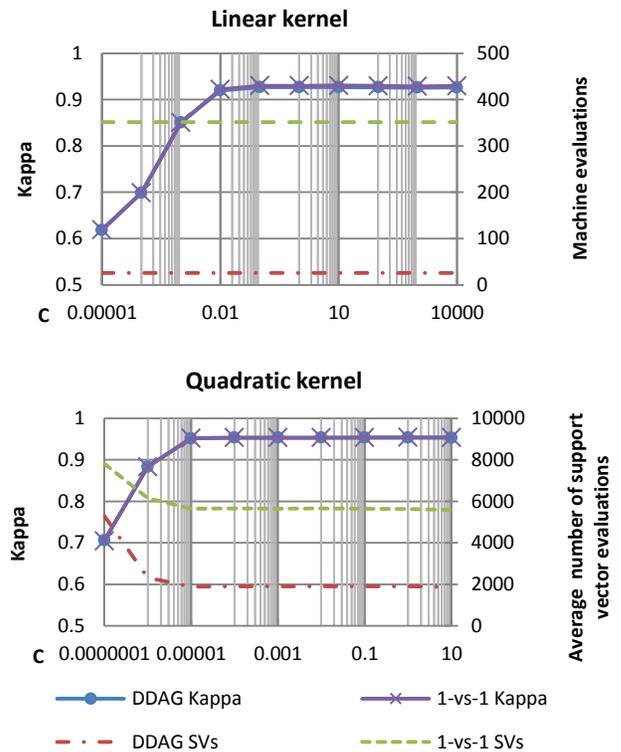

Figure 3. Performance for the Linear and Quadratic polynomial kernels.

The graphs at Figure 3 also provide a comparison for efficiency between the Linear and Quadratic kernels in terms of function and machine evaluations. Since linear machines can be written in compact form, their evaluation speed does not necessarily depend on the number of SVs but rather on the number of machines. Comparisons for accuracy and SVs among selected classifiers are shown in Table 1.

TABLE I.        PERFORMANCE FOR BEST VECTOR MACHINES.

| Kernel | Decision Scheme | Validation | Support Vectors | |
|---|---|---|---|---|
| | | *Kappa ± (0.95 C.I.)* | *Total* | *Average[1]* |
| Linear | DDAG | 0.9268 ± 0.006 | 27,602 | 1,527.34 |
| | 1-vs-1 | 0.9300 ± 0.006 | | 5,483.00 |
| Quadratic | DDAG | 0.9536 ± 0.005 | 37,341 | 1,912.67 |
| | 1-vs-1 | 0.9542 ± 0.004 | | 5,638.00 |
| Gaussian | DDAG | 0.9586 ± 0.004 | 52,220 | 2,518.67 |
| | 1-vs-1 | 0.9583 ± 0.004 | | 6,188.00 |

[1] Average unique evaluations when classifying the set of test instances.

Table 1 displays a dramatic reduction in the average number of SV evaluations actually needed to be computed in the DDAG and voting schemes in comparison to the total number of selected SVs. Because the voting scheme needs to evaluate $c(c-1)/2 = 351$ machines at each decision, its average number of SV evaluations is always equal to the total number of unique vectors found during learning. The DDAG, on the other hand, needs a much smaller, although very variable, number of SVs depending on the evaluation of only $(c-1) = 26$ machines.

Now we start describing results obtained with ANNs. Learned networks were able to achieve similar performance rates of the SVMs, however at a huge training time cost, especially considering the costs involved in running multiple random initializations to ensure a good local minimum. The best values for $\kappa$ were found amid 300~500 neurons, in the same range found in [8]. However it can be seen how the maximum performance obtained by ANNs ($\kappa = 0.9248$) was very similar to the baseline linear SVM ($\kappa = 0.9268$).

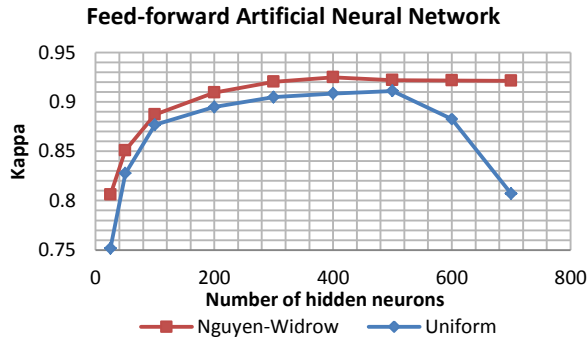

Figure 4.   Performance for ANNs with Resilient Backpropagation

From results, it is possible to see that although Nguyen-Widrow initialization does not affect much the performance of small networks, it had a huge impact on larger ones. For instance, networks with 700 hidden neurons could not learn some classes at all, presenting a zero classification rate for a number of class labels. Networks of the same size trained using Nguyen-Widrow were able to present much more stable results in comparison.

TABLE II.        PERFORMANCE FOR BEST NEURAL NETWORKS.

| Algorithm | Initialization Scheme | Validation | Hidden Neurons |
|---|---|---|---|
| | | *Kappa ± (0.95 C.I.)* | |
| RProp | Uniform (random) | 0.9112 ± 0.006 | 500 |
| RProp | Nguyen-Widrow | 0.9248 ± 0.006 | 400 |

The use of the Nguyen-Widrow initialization proved extremely helpful in reaching good local minima for larger networks. Without the initialization, larger networks would rapidly converge to poor local minima after just a few number of (very time consuming) iterations. The use of the initialization technique seemed to have prevented the often common performance degradation associated with larger-than-necessary networks, as it can be seen in Figure 4. And although not depicted, the use of the heuristic also helped smaller networks attain convergence at much faster rates, with much reduced numbers of iterations.

Furthermore, we can finally note that the best attained SVM ($\hat{\kappa}_{SVM} = 0.9586$, $\widehat{var}(\hat{\kappa}_{SVM}) = 5.098 \times 10^{-6}$) had shown better results than the best ANN found ($\hat{\kappa}_{ANN} = 0.9248$, $\widehat{var}(\hat{\kappa}_{ANN}) = 9.02 \times 10^{-6}$). Considering a $\kappa$ test, those differences are statistically significant under a 0.05 significance level. Those differences would, thus, be unlikely to have occurred only by chance.

## VI.    CONCLUSION

If one would select a classifier based solely on its estimated value for $\kappa$, the clear choice would be to go with a Gaussian or Polynomial SVM classifier. However, those may not be very practical due to the high number of selected SVs which have to be maintained together with the classifier in order to process a new decision. The DDAG decision, on the other hand, provides a considerable increase in evaluation speed, but still requires a possibly highly variable number of vector evaluations to classify a new observation. In applications under which a constant time or regular evaluation speed is desired, this may set DDAGs aside.

Instead, opting for using an ANN also brings many advantages. They perform rather well on this task with a moderate number of hidden neurons, making them very suitable for real-time processing [8]. However, from a learning perspective there can be many disadvantages in this choice. The use of heuristics seems to be crucial to overcome the problem of multiple local minima, and, if training is too time consuming, a proper hyperparameter tuning or the use of more elaborated performance estimators such as k-fold cross-validation can become problematic.

Considering all measured options, the best compromise choice seems to be given by a linear SVM. This model is able to provide evaluation speeds comparable to ANNs; and at the same time, offer the ease of use of a convex learning algorithm. Furthermore, a DDAG based on linear SVMs has a constant evaluation rate, since its evaluation does not depend on the number of support vectors, but rather on the number of classes in the problem. Since a DDAG reduces the evaluation effort from computing 351 constant-time decisions to only 26 constant-time decisions, we attain a highly efficient, moderately performant classifier, suitable to be used in a dynamic gesture recognition system.

ACKNOWLEDGMENT

We would like to express our thanks to CNPq, for partially sponsoring this project; and to Pedroso, who collected the gesture data set used in all experiments. Furthermore, we also would like to thank in advance our referees for sharing valuable comments and views with us.